\documentclass{article}
\usepackage{ijcai24}

\usepackage{times}
\usepackage{soul}
\usepackage{url}
\usepackage[hidelinks]{hyperref}
\usepackage[utf8]{inputenc}
\usepackage[small]{caption}
\usepackage{graphicx}
\usepackage{amsthm}
\usepackage{booktabs}
\usepackage{algorithm}
\usepackage{algorithmic}
\usepackage[switch]{lineno}
\usepackage{algorithm}
\usepackage{algorithmic}
\usepackage{graphicx}
\usepackage{amsmath}
\usepackage{amssymb}
\usepackage{booktabs}
\usepackage{multirow}
\usepackage{array}
\usepackage{gensymb}
\usepackage{color}
\usepackage{mathtools}
\usepackage{subfig}
\newcommand{\etal}{\textit{et al.}}
\usepackage[table]{xcolor}
\definecolor{gray}{RGB}{229, 230, 230}
\usepackage{newfloat}
\usepackage{listings}

\title{Robust Monocular Depth Estimation via Cross-Modal Fusion of \\ RGB and Thermal Images}
\title{Unveiling the Depths: A Multi-Modal Fusion Framework\\ for Challenging Scenarios}

\author{
    Author Name
    \affiliations
    Affiliation
    \emails
    email@example.com
}

\author{
Jialei Xu$^1$
\and
Xianming Liu$^1$,
Junjun Jiang$^{1}$,
Kui Jiang$^{1}$,
Rui Li$^{2}$,
Kai Cheng$^{3}$,
Xiangyang Ji$^4$\\
\affiliations
$^1$Harbin university of Science and Technology, $^2$ETH\\
$^3$University of Science and Technology of China, $^4$Tsinghua University\\
\emails
xujialei@stu.hit.edu.cn,
\{csxm, jiangjunjun,kuijiang\}@hit.edu.cn,
lirui.david@gmail.com,
chengkai21@mail.ustc.edu.cn,
xyji@tsinghua.edu.cn
}

\begin{document}

\maketitle

\begin{abstract}
Monocular depth estimation from RGB images plays a pivotal role in 3D vision. However, its accuracy can deteriorate in challenging environments such as nighttime or adverse weather conditions.
While long-wave infrared cameras
offer stable imaging in such challenging conditions, they are inherently low-resolution, lacking rich texture and semantics as delivered by the RGB image. 
Current methods focus solely on a single modality due to the difficulties to identify and integrate faithful depth cues from both sources. 
To address these issues, this paper presents a novel approach that identifies and integrates dominant cross-modality depth features with a learning-based framework.
Concretely, we independently compute the coarse depth maps with separate networks by fully utilizing the individual depth cues from each modality. 
As the advantageous depth spreads across both modalities, we propose a novel confidence loss steering a confidence predictor network to yield a confidence map specifying latent potential depth areas. With the resulting confidence map, we propose a multi-modal fusion network that fuses the final depth in an end-to-end manner. Harnessing the proposed pipeline, our method demonstrates the ability of robust depth estimation in a variety of difficult scenarios.
Experimental results on the challenging MS$^2$ and ViViD++ datasets demonstrate the effectiveness and robustness of our method.
\end{abstract}
\begin{figure}[th!]
\centering
\includegraphics[width=\linewidth]{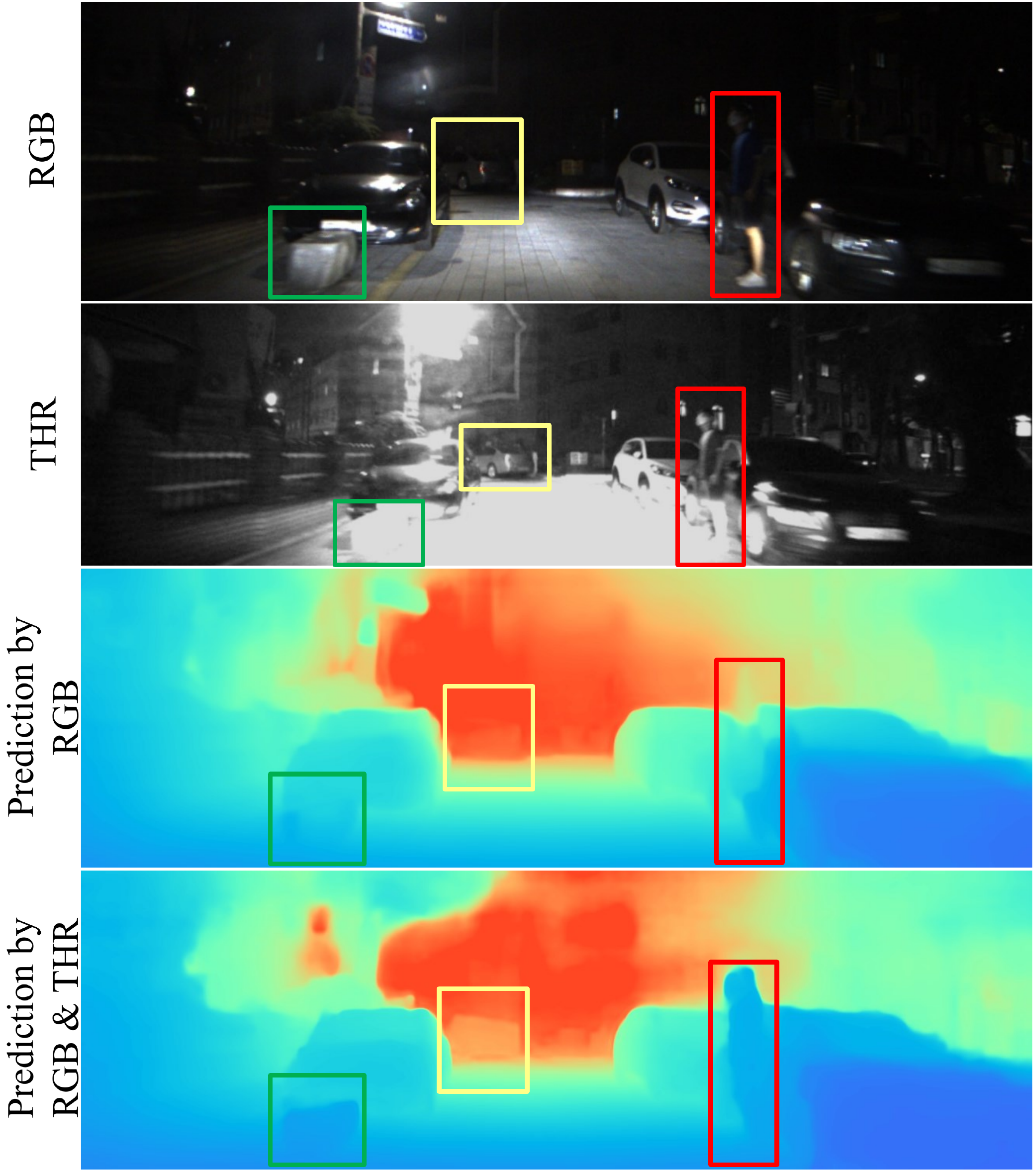}
\vspace{-2em}
\caption{ 
\textbf{A comparison of depth prediction results before and after fusing thermal (THR) images.}
Our method combines depth cues from cross-perspective and cross-modality inputs, yielding high-quality depth in challenging scenes.
}
\label{fig:first_image}
\vspace{-0.5cm}
\end{figure}
\section{Introduction}

\label{sec:intro}


Recent advancements in the field of 3D scene perception have witnessed significant progress in depth estimation from monocular RGB images \cite{bhat2021adabins,yuan2022new,godard2019digging}, with applications proliferating in fields such as autonomous driving and robotics~\cite{godard2019digging,geiger2012we,nister2004visual,he2020review}. The great success of deep neural networks (DNNs) has spurred the proposal of numerous deep learning methodologies tailored to extract precise and dependable 3D data~\cite{bhat2021adabins,yuan2022new,godard2019digging}. However, the environmental sensitivity of RGB cameras, especially under challenging environments like bad weather or nighttime, often compromises the accuracy of RGB-based depth estimation. In contrast, thermal cameras (THR), which capture images by registering the infrared radiation from a target through an optical system, 
are not affected by the visible spectrum. As shown in Fig.~\ref{fig:first_image}, in challenging scenarios such as dark scenes, thermal cameras can clearly capture objects that RGB camera struggles with in absence of the visible lights, showcasing a more consistent performance under such complex conditions. 


When developing depth prediction methods based on thermal images,
they typically take thermal images as the only input source, with the mirrored architectures initially designed for RGB images.
This is especially noticeable in recent methods, ranging from the stereo depth estimation~\cite{shin2023deep} to self-supervised depth estimation techniques~\cite{shin2021self,shin2022maximizing}. However, estimating depth from only the thermal image can also lead to suboptimal results. Concretely, thermal images typically possess lower resolutions compared to RGB images, making high-fidelity 3D perception more challenging. 
Additionally, thermal images mainly reflect the heat information of the scene, which inherently lack the capability to express fine textures and semantics as delivered by RGB images.
Hence, 
it is crucial to combine the depth cues of the two modalities and make their advantages complement each other for robust depth estimation.
Existing depth estimation methods, whether focus on thermal~\cite{shin2019sparse,shin2023deep} or RGB images~\cite{Yin_2019_ICCV,yin2023metric3d,yuan2022new}, tend to operate independently, thus overlooking the potential benefits of combining the both. 

Applying existing multi-modal fusion approaches directly to depth estimation tasks presents substantial challenges.
The depth-related advantages offered by both modalities are irregularly  dispersed throughout the scene. Screening information that is beneficial to depth prediction plays an important role in multi-modal fusion.
Traditional approaches, whether based on manual crafting or inductive biases, struggle to accurately identify and fuse these faithful depth cues. 
The process of precisely locating and effectively integrating these cues stands as a technical challenge, necessitating dedicated research to develop accurate methods.

Building on these insights, we advocate for the utilization of thermal images in tandem with RGB images for robust monocular depth estimation in challenging environments. Our approach endeavors to harmonize the strengths of both imaging methods, ensuring accurate 3D perception in complex environments. To fully adopt the advantages of each modality, our method begins with two depth estimation networks to approximate the coarse depth maps for both RGB and thermal images respectively. 
Aiming to identify and fuse the faithful depth cues from aligned depth maps, we propose a confidence predictor network associated with a novel confidence loss to locate informative depth areas that each modality is potentially confident at. 
The resulting confidence map guides the following fusion network, which amalgamates the advantages of both modalities to refine the coarse depth for the final depth prediction. 
Combining these contributions, our method produces high-quality depth in challenging scenes.
As a result, our method achieves state-of-the-art performance across current benchmarks.
We summarize our contributions in the following:





\begin{itemize}
\item The proposed approach is the first to utilize thermal images in tandem with RGB image for monocular depth estimation, and we discover and prove that different modalities have their own advantages and disadvantages in depth estimation.
\item
The proposed confidence predictor network, confidence loss and  fusion network are able to identify and fuse the advantages of both modalities to refine the coarse depth.


\item The proposed approach is able to predict depth in multiple complex scenarios and outperforms the results of a single modality, as demonstrated by experimental results on popular multi-modal datasets such as MS$^2$~\cite{shin2023deep} and ViViD++~\cite{lee2022vivid++}.
\end{itemize}

\section{Related Work}

\subsection{Monocular Depth Estimation from RGB}
The monocular depth estimation task is to recover the three-dimensional information of the scene using a single RGB image.  Eigen~\etal propose one of the earliest works in convolutional-based depth estimation using an end-to-end deep network to regress the depth directly from a single image. A lot of subsequent works 
~\cite{lee2019big,song2021monocular,ramamonjisoa2019sharpnet} follow this pipeline to improve the network structure and continuously improve the accuracy. Instead of treating depth prediction as a regression problem, ~\cite{bhat2021adabins,xu2021weakly} consider the depth estimation problem as a classification problem. They classify each pixel by discretizing the depth to complete the depth estimation task, and achieved excellent results. 
Yin~\etal~\cite{Yin_2019_ICCV} demonstrate the importance of the high-order 3D geometric constraints for depth prediction.
At present, one of the difficulties in the depth estimation from cameras is how to improve the robustness and accuracy of complex scenes, such as night, rainy days, etc. ~\cite{wang2021regularizing} propose priors-based regularization to learn distribution knowledge from unpaired depth maps and prevent model from being incorrectly trained in the night.
Li~\etal~\cite{li2021single} distinguish real and fake on the generated depth image using a discriminator, and enhances the performance of the generator in the dark scene. 

\begin{figure*}[t!]
\centering
\includegraphics[width=18.0cm]{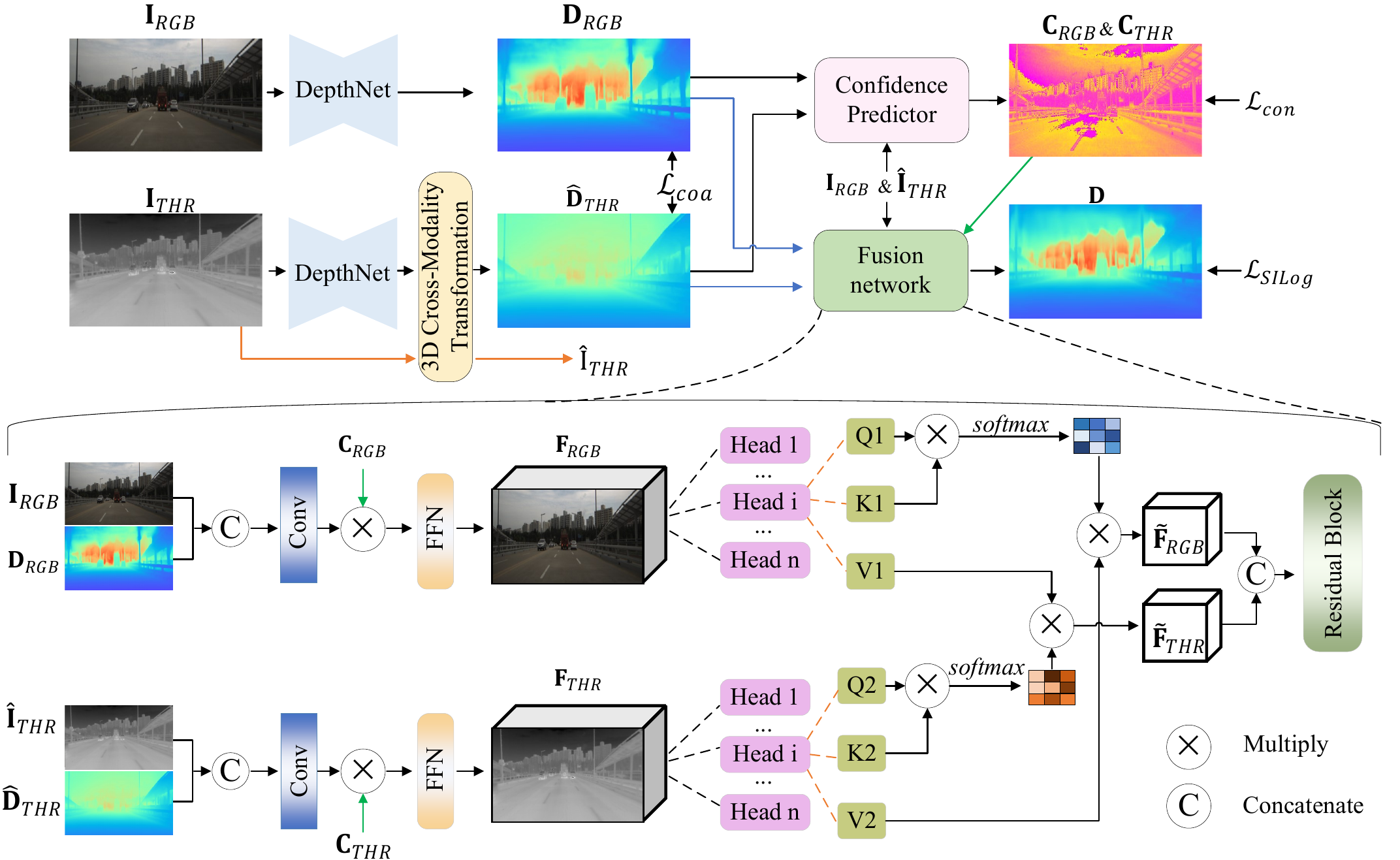}
\vspace{-1em}
\caption{\textbf{Overall pipeline of our proposed depth estimation via cross-modal fusion of RGB and thermal images (THR).} We first employ two distinct depth networks to estimate the coarse depth map of each modality. Using 3D cross-modal transformation, we ensure that the information of the two modalities can be aligned in the same perspective. The confidence map (\emph{i.e.}, $\mathbf{C}_{RGB}$ and $\mathbf{C}_{THR}$) calculated by the confidence predictor network can identify which modality coarse depth can more realistically reflect the 3D scene. With the guidance of confidence maps, the fusion network fuses the advantages of both modalities to refine the coarse depth for the final depth prediction.
}
\vspace{-1em}
 \label{fig:pipline}
\end{figure*}

\subsection{Depth Estimation From Thermal Image}
The infrared spectrum band reflects the heat information of the environment, and its images (\emph{i.e.} thermal image) are very robust to the environment and weather compared to the visible spectrum band.
At present, there are a lot of work using thermal image for 3D vision such as visual odometry~\cite{delaune2019thermal,khattak2020keyframe} and SLAM ~\cite{shin2019sparse}, and achieve good robustness. In the depth estimation task, most of the works focus on self-supervised monocular thermal camera to predict depth maps.
They all follow the same algorithmic processes applied to RGB images. That is by calculating the relative attitude of a camera moving in a video sequence, and the predicted depth information to construct geometric constraint relationship as the training network supervision information. Shin~\cite{shin2023deep} first propose using stereo thermal cameras to estimate disparity to calculate depth maps, and propose a unified depth network that effectively bridges monocular depth and stereo depth tasks from a conditional random field perspective.  

\subsection{Multimodal Tasks in Perception Systems}
Thanks to the stability of infrared cameras in harsh environments, many works have attempted to use RGB and THR modes together to enhance accuracy and robustness, such as scene parsing~\cite{zhou2022edge}, semantic segmentation~\cite{lai2023mefnet}, image enhancement~\cite{shopovska2019deep}, and object detection~\cite{el2023enhanced,zhou2023lsnet}.
Compared to a single RGB image, more information from thermal image (THR) can bring richer scene information, which naturally improves the accuracy of the target task. 
These methods have in common a complex network structure to fuse the features of the two modalities for better performance. However, we find that each modality has its own advantages and disadvantages. Locating and fusing the dominant information, discarding the information with large error are more helpful to the effect of fusion.
Another characteristic of existing multi-modal tasks~\cite{wang2018rgb,shivakumar2019pst900} is that they rarely consider the perspective changes between different modalities. 
We adopt the 3D transformation commonly used in multi view stereo~\cite{gu2020cascade} to address the issue of viewpoint differences. This is not an innovative scheme, but it is important for solving multi-modal work.

\section{Method}

\subsection{Problem Formulation}
\label{sec:problem formulation}
Given data collections from an RGB and a thermal camera, two simultaneous outputs can be obtained: an RGB image, denoted as $\mathbf{I}_{RGB}$, and a thermal image, denoted as $\mathbf{I}_{THR}$. The intrinsic parameters for both cameras are respectively described as $\mathbf{K}_{RGB}$ and $\mathbf{K}_{THR}$. Additionally, the extrinsic parameters $\mathbf{E}$ derived from the dataset are represented as ${\mathbf{E}} = \begin{psmallmatrix}\mathbf{R} & \mathbf{t}\\ \mathbf{0} & \mathbf{1}\end{psmallmatrix} \in \text{SE(3)}$, where $\mathbf{R}$ and $\mathbf{t}$ correspond to the rotation and translation parameters between the two cameras.

The primary objective of this study is to formulate a depth prediction model $f(\cdot|\boldsymbol{\theta})$, which predicts the depth map $\mathbf{D}$ from the camera perspective using $\mathbf{I}_{RGB}$ in conjunction with $\mathbf{I}_{THR}$ as inputs.
We focus on the fully-supervised task, where the data procured from the depth sensor $\mathbf{D}_{gt}$ serves as the ground truth during the network training phase.

\begin{figure}[t!]
\centering
\subfloat{
\parbox[t]{2mm}{\rotatebox{90}{\small ~~~~~~~~~~~~~~}}
\includegraphics[width=0.97\linewidth]{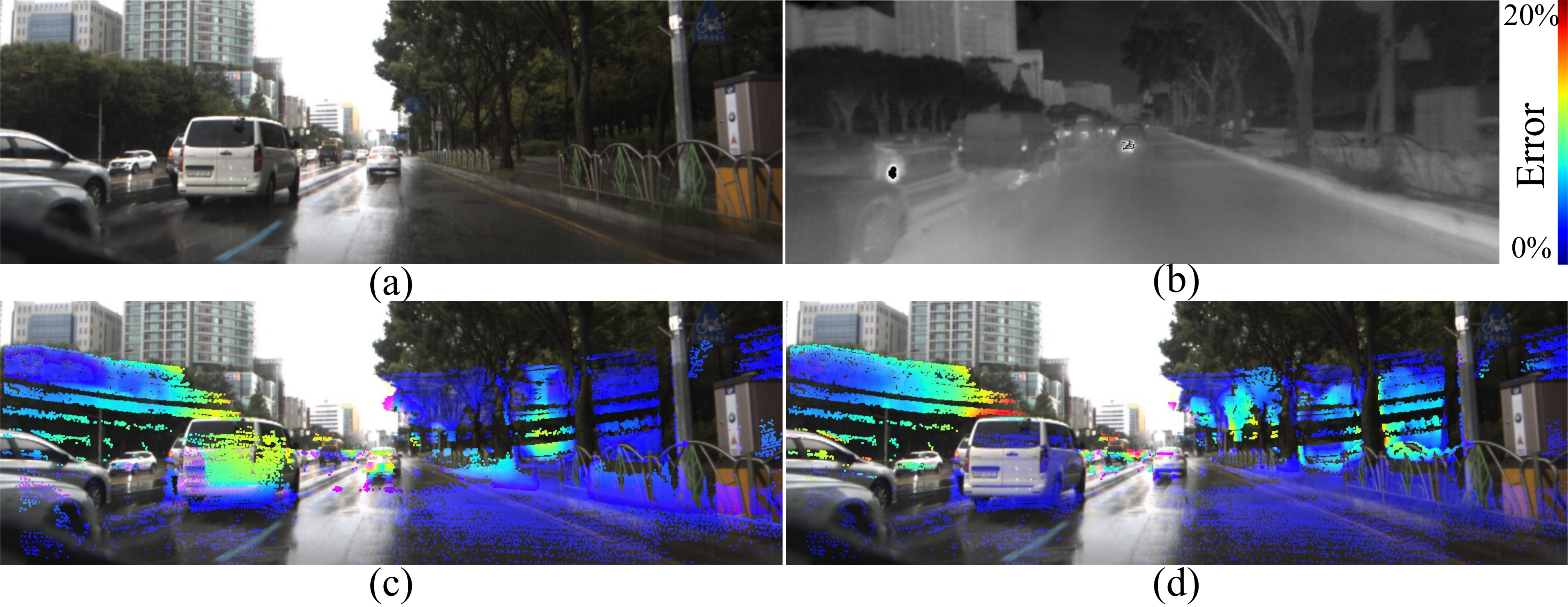}
}
\vspace{-1em}
\caption{\textbf{Error maps for predicting depth from RGB and thermal images respectively.} (a) RGB (b) THR (c) Error map of depth prediction for RGB image (d) Error image of depth prediction for thermal image. The error map represents the relative error between the predicted depth and the ground-truth. We specify values for different color response errors and overlay them on RGB images.
}
\label{fig:error map}
\end{figure}

\subsection{Coarse Depth from One Modal}
\label{sec:coarse depth}
In multi-modal depth estimation, since each modality delivers depth cues that may benefit each other, 
herein we employ two separate neural networks to estimate the depth maps from $\mathbf{I}_{RGB}$ and $\mathbf{I}_{THR}$, yielding $\mathbf{D}_{RGB}$ and $\mathbf{D}_{THR}$ respectively. 
We adopt the basic convolutional model~\cite{godard2019digging} as the depth network, which is a standard regression model and has been applied in many works~\cite{guizilini2022full,xu2022multi}. 
The depth estimation networks employed for both modalities operate independently.
Contrary to the approach presented by~\cite{shin2023self}, our networks do not  share parameters, nor do we impose any similarity constraints on their features. 
This distinction holds significance as each modality possesses distinct advantages, and it is crucial to avoid confusion by keeping their individual features.

\noindent\textbf{3D Cross-Modal Transformation.}
To accurately align the information of the two modalities, we adopt 3D coordinate transformation~\cite{gu2020cascade} as a 3D cross-modal transformation module. 
This alignment is achieved through a synergistic utilization of intrinsic parameters, $\mathbf{K}_{RGB}$ and $\mathbf{K}_{THR}$, complemented by the extrinsic parameters represented as ${\mathbf{E}} = \begin{psmallmatrix}\mathbf{R} & \mathbf{t}\\ \mathbf{0} & \mathbf{1}\end{psmallmatrix}$. 
Moreover, by incorporating the coarse depth, $\mathbf{D}_{THR}$, we are able to devise a pixel-warping operation to transition from the thermal to the RGB image as follows:

\begin{equation}
\hat{\mathbf{p}}_{RGB} = \pi \big(\mathbf{R}~\phi (\mathbf{p}_{THR}, \mathbf{D}_{THR},{\mathbf{K}_{THR}}) + {\mathbf{t}},{\mathbf{K}_{RGB}}\big),
\label{eq:warp_spatial}
\end{equation}
where $\phi(\mathbf{p}, \mathbf{d}, \mathbf{K}) = \mathbf{P}$ refers to the back-projection of a pixel in homogeneous coordinates $\mathbf{p}$ to a 3D point $\mathbf{P}$ for a given estimated depth $\mathbf{d}$. $\hat{\mathbf{p}} = \pi(\mathbf{P},{\mathbf{K}})$ denotes the projection of a 3D point back onto the image plane, where the subscripts $RGB$ and $THR$ denote the RGB and thermal images. Using the linear interpolation operation based on the coordinate $\hat{\mathbf{p}}_{RGB}$, we can calculate the depth predicted by the thermal image in the perspective of the RGB camera, denoted by $\mathbf{\hat{D}}_{THR}$.

\noindent\textbf{Coarse Depth Loss.}
It is crucial to ensure the quality of the coarse depth map. High-quality depth results can improve the accuracy of the 3D cross-modal transformation and provide a favorable basis for subsequent confidence predictor network and fusion networks. To this end, we utilize the ground-truth $\mathbf{D}_{gt}$ to supervise the predicted coarse depth using the $L_1$ loss function. We define the coarse loss $\mathcal{L}_{coa}$ as follows:

\begin{equation}
\mathcal{L}_{coa} = L_{1}(\mathbf{D}_{RGB},\mathbf{D}_{gt}) + L_{1}(\mathbf{\hat{D}}_{THR},\mathbf{D}_{gt}).
\end{equation}
We use $L_{coa}$ to train the depth estimation networks.

\subsection{Confidence Map of RGB and Thermal Images}
\label{sec:confidence map}

We visualize the absolute relative error (Abs Rel) maps between the predicted coarse result (\emph{i.e.}, $\mathbf{D}_{RGB}$ and $\mathbf{\hat{D}}_{THR}$) and the ground-truth. As shown in Fig.~\ref{fig:error map},
in the same scenario, the prediction results of different modalities are very different and each modal has unique advantages and disadvantages. For example, RGB cameras show better depth estimation results in areas with rich textures, but it struggles with reflective areas.
Thermal images exhibit stable imaging under different lighting conditions, but are limited to low resolution and have a decreased ability to perceive object details.
Our overarching goal is to filter out the less accurate regions and amalgamate the areas of higher precision to formulate the final depth estimation results. 
Consequently, it is imperative that the neural network is adept at discerning the accuracy of prediction results from each modality. 
This ability of identification serves as a pivotal precursor for the efficacious fusion of information from the two modalities.

\noindent\textbf{Confidence Predictor Network.} 
In pursuit of discerning which modality yields more accurate results, we architect a specialized confidence predictor network. 
Inputs for this network include $\mathbf{I}_{RGB}$, $\mathbf{I}_{THR}$, $\mathbf{D}_{RGB}$ and $\mathbf{\hat{D}}_{THR}$.
Prior to computational processing within the network,  Eq.~\ref{eq:warp_spatial} is employed to re-project $\mathbf{I}_{THR}$ from the thermal camera's perspective to that of the RGB camera, denoted as $\mathbf{\hat{I}}_{THR}$.
This transformation ensures consistent perspectives across all input images, aligning their pixel information accordingly.
The output of confidence predictor network is a confidence map for RGB, denoted as $\mathbf{C}_{RGB}$ and dimensioned as $\mathbb{R}^{1\times H \times W}$, with $H$ and $W$ representing the height and width of RGB image, respectively. 
The confidence map $\mathbf{C}_{RGB}$ serves as an indicative measure of the proximity between $\mathbf{D}_{RGB}$ and the ground-truth, whose values range between 0 and 1. 
A value approaching 1 implies that the depth estimated from the RGB image closely aligns with the ground-truth, while a value of 0 suggests a completely erroneous prediction.  This confidence map delineates the relative accuracies between the results from both modalities (\emph{i.e.}, $\mathbf{D}_{RGB}$ and $\mathbf{D}_{THR}$) in relation to the ground-truth. Therefore, the confidence map for the thermal modality can be subsequently computed as:

\begin{equation}
\mathbf{C}_{THR} = \mathbf{1} - \mathbf{C}_{RGB}.
\end{equation}
We employ an UNet architecture with the Resnet~\cite{he2016deep} backbone as the confidence predictor network, and adopt basic convolutional blocks as the decoder.

\noindent\textbf{Confidence Loss.} 
In our endeavors to bolster the confidence predictor network's precision in gauging the proximity between the predicted values and the ground-truth, we introduce a specialized confidence loss, denoted as $\mathcal{L}_{con}$. 
To commence, we mathematically define the error associated with the predicted depths from both modalities in relation to the ground-truth $\mathbf{{D}}_{gt}$ as articulated below:

\begin{equation}
\begin{split} 
\mathbf{E}_{RGB} = \|\mathbf{{D}}_{gt} - \mathbf{\mathbf{D}}_{RGB}\|_1,\\
\mathbf{E}_{THR} = \|\mathbf{{D}}_{gt} - \mathbf{\mathbf{\hat{D}}}_{THR}\|_1.
\label{eq:confidence loss} 
\end{split}
\end{equation}
Drawing upon the inherent implication of the confidence map, the confidence loss, $\mathcal{L}_{con}$, is formalized as:

\begin{equation}
\begin{split} 
\mathcal{L}_{con} &= \|\frac{exp({\mathbf{E}_{THR}})} {exp({\mathbf{E}_{RGB})} +  exp({\mathbf{E}_{THR}})} - \mathbf{C}_{RGB}\|_1 \\
&+\|\frac{exp({\mathbf{E}_{RGB}})} {exp({\mathbf{E}_{RGB}}) +  exp({\mathbf{E}_{THR}})} - \mathbf{C}_{THR}\|_1,
\label{eq:confidence loss}
\end{split}
\end{equation}
where $exp()$ represents the exponential function based on the natural logarithm's foundation.
In the first term of the loss function $\mathcal{L}_{con}$ , when the error of RGB is lower than the error of THR, the confidence score of RGB obtains a higher value. The second term achieves the same goal for THR confidence. The purpose of exponential function in loss function is to expand the proportion of the dominant modality in the fusion network and avoid losing the inferior mode completely (exp(0)=1).
The defined $\mathcal{L}_{con}$ is instrumental in equipping the confidence predictor network with the capability to identify the respective accuracy of $\mathbf{D}_{RGB}$ and $\mathbf{D}_{THR}$. Consequently, this confidence loss term, $\mathcal{L}_{con}$, is integrated into the overall loss function during the network's training phase.

\subsection{Cross-modal Information Fusion}

\label{sec:fusion}
In order to achieve cross-modal information fusion of 3D structural information expressed in RGB images and thermal images, we design the fusion network. This network adopts images of each modality (\emph{i.e.}, $\mathbf{I}_{RGB}$ and $\mathbf{\hat{I}}_{THR}$) and the predicted depth coarse maps (\emph{i.e.}, $\mathbf{D}_{RGB}$ and $\mathbf{D}_{THR}$)  as part of the input.
The guidance of confidence maps, $\mathbf{C}_{RGB}$ and $\mathbf{C}_{THR}$, is incorporated to 
help the fusion network select the dominant information in each modality and discard the error results.
The ultimate objective is to synergistically combine the inherent advantages of the two modalities, resulting in the final depth estimation denoted as $\mathbf{D}$.


The schematic diagram of the fusion network is shown in Fig.~\ref{fig:pipline}. Specifically, the fusion network adopts a dual-branch structure, and each branch first independently calculates the information of one modality. Taking RGB images as an example, we first concatenate (\emph{cat}) the $\mathbf{I}_{RGB}$ and $\mathbf{D}_{RGB}$ along the channel dimension, which is then followed by a convolution layer (\emph{conv}) to extract low-level features. 
Here, the integration of confidence maps is achieved through element-wise multiplication operations (\emph{mul}) between the feature and $\mathbf{C}_{RGB}$, subsequently producing feature maps $\mathbf{F}_{RGB}$ through a feedforward network ($FFN$), as depicted by the following:

\begin{equation}
\begin{split} 
\mathbf{F}_{RGB} &= FFN(conv(cat(\mathbf{I}_{RGB},\mathbf{D}_{RGB}))\cdot \mathbf{C}_{RGB}),\\
\mathbf{F}_{THR} &= FFN(conv(cat(\mathbf{\hat{I}}_{THR},\mathbf{D}_{THR}))\cdot \mathbf{C}_{THR}).
\end{split}
\end{equation}

\noindent\textbf{Multi-head Fusion Attention.}
$\mathbf{F}_{RGB}$ and $\mathbf{F}_{THR}$ have corresponding modal advantage feature information on different branches. To fuse them effectively, we introduce a multi-head fusion attention module, as depicted in Fig.~\ref{fig:pipline}. 
Given $\mathbf{F}_{RGB}$ and $\mathbf{F}_{THR}$ as inputs, the attention module ($Att$) can be expressed as follows:

\begin{equation}
\begin{split} 
\mathbf{\widetilde{F}}_{RGB}=Att(Q_1,K_1,V2) = softmax({{Q_1}{K_1}^{T})V_2},\\
\mathbf{\widetilde{F}}_{THR}=Att(Q_2,K_2,V1) = softmax({{Q_2}{K_2}^{T})V_1},
\end{split}
\end{equation}
where $Q$ denotes the query, ${K}$ stands for the key and $\emph{V}$ is the value. $\mathbf{\widetilde{F}}_{RGB}$ and $\mathbf{\widetilde{F}}_{THR}$ are the two output features of the multi-head fusion attention module. $softmax$ denotes the softmax function. We concat $\mathbf{\widetilde{F}}_{RGB}$ and $\mathbf{\widetilde{F}}_{THR}$ along the channel, which is input to the residual block. The residual block contains convolutional layers, normalization layers, Leaky ReLU function~\cite{xu2015empirical}, and interpolation operation. The output is the final target depth map $\mathbf{D}$. 
The result of the fusion is more accurate than any single modality prediction, and we analyze the effect of the model in detail in the experiments.

\subsection{Training Loss.}
Following previous works~\cite{yuan2022new,bhat2021adabins,lee2019big}, we use the simple and common Scale-Invariant Logarithmic loss between predicted depth and ground-truth to supervise the training:

\begin{equation}
    \Delta \mathbf{D}_i = \log {{\mathbf{D}}_{gt}}_i - \log \mathbf{D}_i,
\end{equation}
where $\mathbf{D}_{gt}$ is the ground-truth depth value and $\mathbf{D}$ is the predicted depth at pixel $i$.
Then for $K$ pixels with valid depth values in an image, the scale-invariant logarithmic loss is computed as:

\begin{equation}
    \mathcal{L}_{SILog} = \sqrt{\frac{1}{K} \sum_i\Delta {\mathbf{D}_i}^2 - \frac{\lambda}{K^2} (\sum_i \Delta {\mathbf{D}_i}^2)},
\end{equation}
where $\lambda$ is a variance minimizing factor, and is set to $0.5$ in our experiments.


Combined with the two losses we proposed, the overall loss is computed as:

\begin{equation}
    \mathcal{L} = \mathcal{L}_{coa} + \beta \mathcal{L}_{con} + \gamma \mathcal{L}_{SILog},
\end{equation}
where $\beta$ and $\gamma$ are weight parameters, and set to 0.8 and 0.65 in our experiments.

\begin{table*}[t!]
\begin{center}
\renewcommand\arraystretch{0.95}
\setlength\tabcolsep{5pt} 
\small
\begin{tabular}{c|c|cccc|ccc}
  \hline
\multirow{2}{*}{\textbf{Methods}}  &\multirow{2}{*}{\textbf{Modality}} &
\multicolumn{4}{c|}{Error $\downarrow$}&
\multicolumn{3}{c}{Accuracy $\uparrow$}\\

\cline{3-9}
&&AbsRel&SqRel&RMSE&RMSElog&$\delta<1.25$&$\delta<1.25^{2}$&$\delta<1.25^{3}$\\
\hline
\hline
Adabins~\cite{bhat2021adabins} & RGB & 0.129  & 1.108 & 5.462  & 0.236 & 0.845 & 0.940 & 0.978 \\
NeWCRFs~\cite{yuan2022new}&RGB & 0.124  & 0.997 & 5.193  & 0.211 & 0.831 & 0.928 & 0.974 \\
MIM~\cite{xie2023revealing}&RGB& 0.123 & 1.009 & 5.276  & 0.223 & 0.859 & 0.941 & 0.978 \\
\hline
MTVUMCL~\cite{shin2021self}&THR &0.153 &	1.358 &	6.031   & 0.274 & 0.818 & 0.920 & 0.964\\
MSFT~\cite{shin2022maximizing}&THR&0.126&1.274&	5.740   & 0.261 & 0.824 & 0.929 & 0.969\\
DDETI~\cite{shin2023deep}& THR &	0.123 &	1.103 &	5.134   & 0.208 & 0.869 & 0.948 & 0.975\\
\hline
MURF ~\cite{xu2023murf} &RGB \& THR&	0.118 & 	1.015 &     5.123& 	0.209 & 	0.870 & 	0.951 & 	0.978\\
MCT~\cite{wang2023mct} & RGB \& THR & \underline{0.117} & \underline{1.006} & \underline{5.116} & \underline{0.205} & \underline{0.873} & \underline{0.953} & \underline{0.978}
\\
Ours & RGB \& THR               &\textbf{0.106}&\textbf{0.959}&\textbf{5.003}&\textbf{0.194}&\textbf{0.882}&\textbf{0.960}&\textbf{0.984}\\
\hline
\end{tabular}
\vspace{-1em}
\caption{\protect\label{tab:result-ms2} \protect\textbf{Quantitative evaluations of depth estimation results  on the MS$^2$ dataset~\protect\cite{shin2023deep}.} 
}
\vspace{-0.5em}
\end{center}
\end{table*}

\begin{table*}[t!]
\begin{center}
\renewcommand\arraystretch{0.99}
\setlength\tabcolsep{8pt} 
\small
\begin{tabular}{c|c|c|cccc|ccc}
  \hline
\multirow{2}{*}{\textbf{TestSet}}  &\multirow{2}{*}{\textbf{Modality}} &
\multirow{2}{*}{\textbf{Method}}&
\multicolumn{4}{c|}{Error $\downarrow$}&
\multicolumn{3}{c}{Accuracy $\uparrow$}\\

\cline{4-10}
&&&AbsREL&SqRel&RMSE&RMSElog&$\delta<1.25$&$\delta<1.25^{2}$&$\delta<1.25^{3}$\\
\hline
\hline
\multirow{3}{*}{Day} &\multirow{2}{*}{{Baseline}} & RGB & 0.097  & 0.947 & 4.938  & 0.200 & 0.888 & 0.959 & 0.988 \\
 && THR & 0.107  & 0.911 & 5.174  & 0.203 & 0.844 & 0.958 & 0.985 \\
 &\cellcolor{gray}\multirow{1}{*}{{Ours}}& \cellcolor{gray}RGB \& THR & \cellcolor{gray}\textbf{0.093}  & \cellcolor{gray}\textbf{0.893} & \cellcolor{gray}\textbf{4.864}  & \cellcolor{gray}\textbf{0.187} & \cellcolor{gray}\textbf{0.899} & \cellcolor{gray}\textbf{0.968} & \cellcolor{gray}\textbf{0.990} \\

\hline

 \multirow{3}{*}{Night}&\multirow{2}{*}{{Baseline}}&RGB   & 0.137  & 1.126 & 5.460  & 0.229 & 0.814 & 0.940 & 0.973 \\
&& THR & 0.104  & 0.935 & 4.991  & 0.200 & 0.861 & 0.966 & 0.987 \\
&\cellcolor{gray}\multirow{1}{*}{{Ours}}& \cellcolor{gray} RGB \& THR & \cellcolor{gray}\textbf{0.096}  &\cellcolor{gray} \textbf{0.925} & \cellcolor{gray}\textbf{4.859}  & \cellcolor{gray}\textbf{0.180} & \cellcolor{gray}\textbf{0.898} & \cellcolor{gray}\textbf{0.969} & \cellcolor{gray}\textbf{0.988} \\

\hline

\multirow{3}{*}{Rain}&\multirow{2}{*}{{Baseline}}   & RGB & 0.139  & 1.216 & 5.494  & 0.245& 0.820 & 0.928 & 0.971 \\
&& THR & 0.145  & 1.231 & 5.848  & 0.237 & 0.785 & 0.936 & 0.964 \\ 
&\cellcolor{gray}\multirow{1}{*}{{Ours}}& \cellcolor{gray}RGB \& THR &\cellcolor{gray} \textbf{0.134}  &\cellcolor{gray} \textbf{1.119} & \cellcolor{gray}\textbf{5.453}  & \cellcolor{gray}\textbf{0.225} & \cellcolor{gray}\textbf{0.839} & \cellcolor{gray}\textbf{0.939} & \cellcolor{gray}\textbf{0.973} \\
\hline

\end{tabular}
\vspace{-1em}
\caption{\protect\label{tab:different weather}\protect\textbf{Quantitative evaluations of our method  under diverse scenarios on the MS$^2$ dataset~\protect\cite{shin2023deep}.} 
MS$^2$ dataset contains three subsets: Day, Night, Rain. The baseline of RGB or THR denote the coarse depth $\mathbf{D}_{RGB}$ or $\mathbf{D}_{THR}$ in Sec.~3.2.
}
\end{center}
\vspace{-1em}
\end{table*}
\begin{table*}[t!]
\begin{center}
\renewcommand\arraystretch{0.99}
\setlength\tabcolsep{7.0pt} 
\small
\begin{tabular}{c|c|cccc|ccc}
  \hline
\multirow{2}{*}{\textbf{Methods}}  &\multirow{2}{*}{\textbf{Modality}} &
\multicolumn{4}{c|}{Error $\downarrow$}&
\multicolumn{3}{c}{Accuracy $\uparrow$}\\

\cline{3-9}
&&AbsREL&SqRel&RMSE&RMSElog&$\delta<1.25$&$\delta<1.25^{2}$&$\delta<1.25^{3}$\\
\hline
\hline
Adabins~\cite{bhat2021adabins} &RGB& 0.315  & 0.512 & 0.699  & 0.302 & 0.671 & 0.902 & 0.980 \\
MIM~\cite{xie2023revealing}& RGB  & 0.296  & 0.498 & 0.671  & 0.297 & 0.688 & 0.905 & 0.981 \\
\hline
MTVUMCL~\cite{shin2021self} & THR    & 0.071  & 0.049 & 0.296  & 0.114 & 0.942 & 0.974 & 0.991 \\
MSFT~\cite{shin2022maximizing}& THR & 0.067  & 0.046 & 0.278  & 0.117 & 0.948 & 0.980 & 0.992 \\
\hline
MURF~\cite{xu2023murf} & RGB \& THR & 0.064 & \underline{0.041} & 0.277 & 0.112 & 0.949 & 0.981 & \underline{0.993}\\
MCT~\cite{wang2023mct} & RGB \& THR &  \underline{0.062} & 0.043 & \underline{0.272} & \underline{0.111} & \underline{0.951} & \underline{0.983} & 0.992\\
Ours & RGB \& THR & \textbf{0.054}  & \textbf{0.035} &\textbf{0.267}  & \textbf{0.104} & \textbf{0.956} & \textbf{0.988} & \textbf{0.994} \\
\hline

\end{tabular}
\vspace{-1em}
\caption{\protect\textbf{\protect\label{tab:result-ViViD++} Quantitative evaluations of our method on the ViViD++ dataset~\protect\cite{lee2022vivid++}.} 
We show SOTA methods for depth estimation using RGB or thermal images alone as a reference.}
\end{center}
\vspace{-1em}
\end{table*}

\begin{figure*}[t!]
\centering

\subfloat{
\parbox[t]{2mm}{\rotatebox{90}{\small ~~~~~~~~~(a)}}
\includegraphics[width=0.33\linewidth]{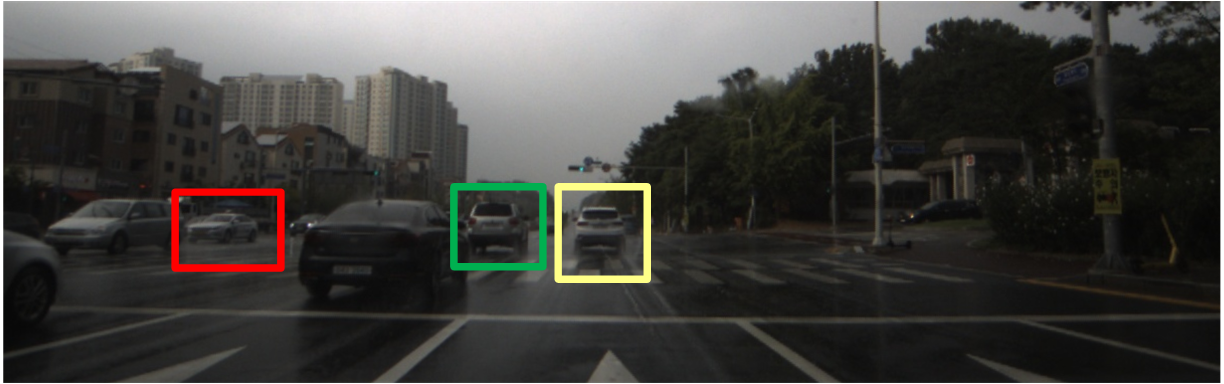}
\includegraphics[width=0.33\linewidth]{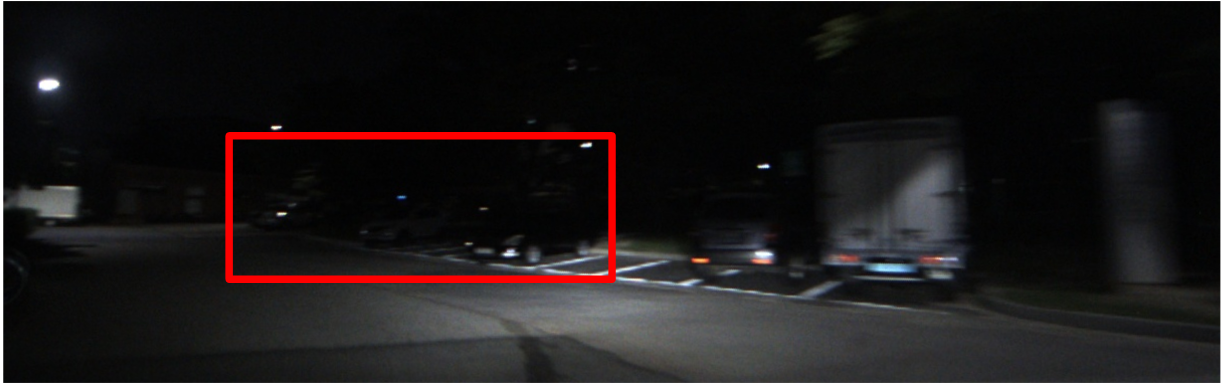}
\includegraphics[width=0.33\linewidth]{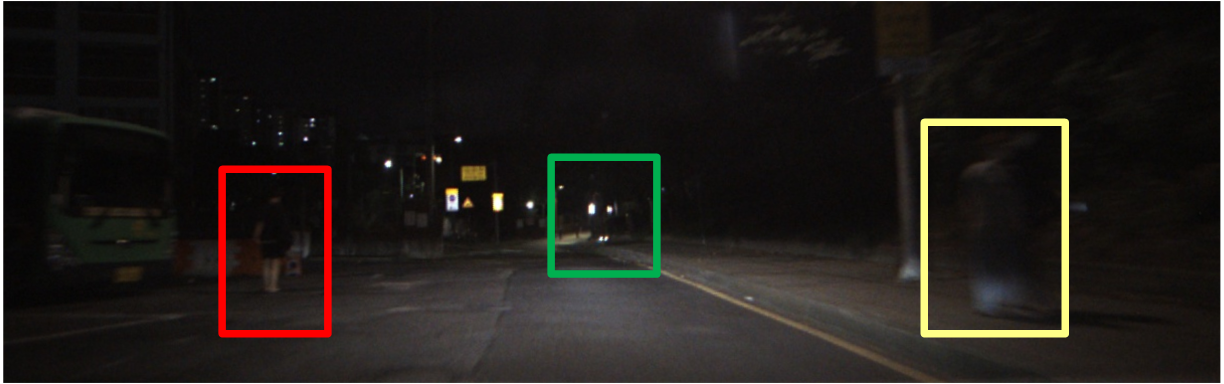}
}\\
\vspace{-1em}
\subfloat{
\parbox[t]{2mm}{\rotatebox{90}{\small ~~~~~~~~~(b)}}
\includegraphics[width=0.33\linewidth,height=1.8cm]{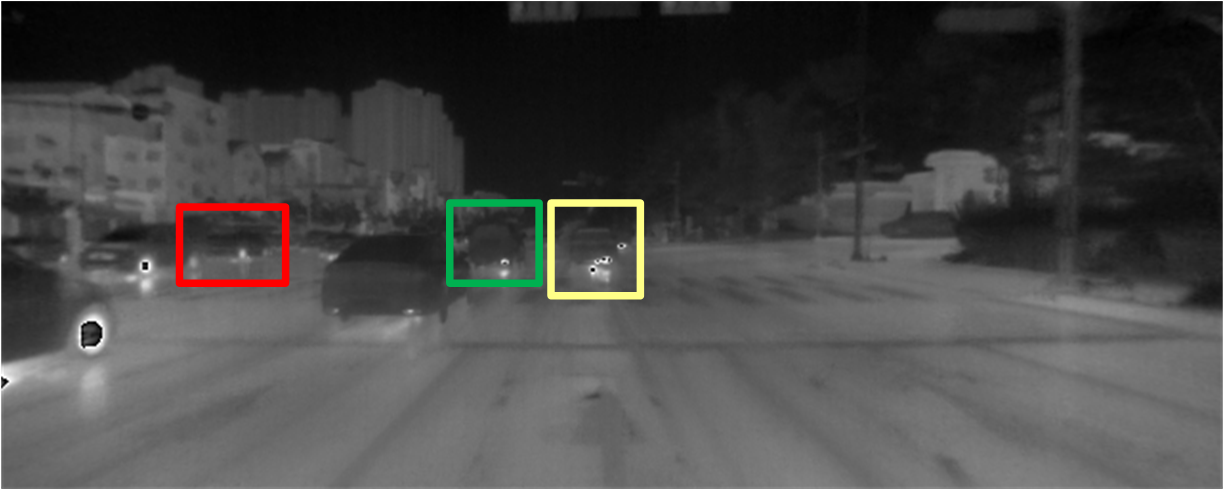}
\includegraphics[width=0.33\linewidth,height=1.8cm]{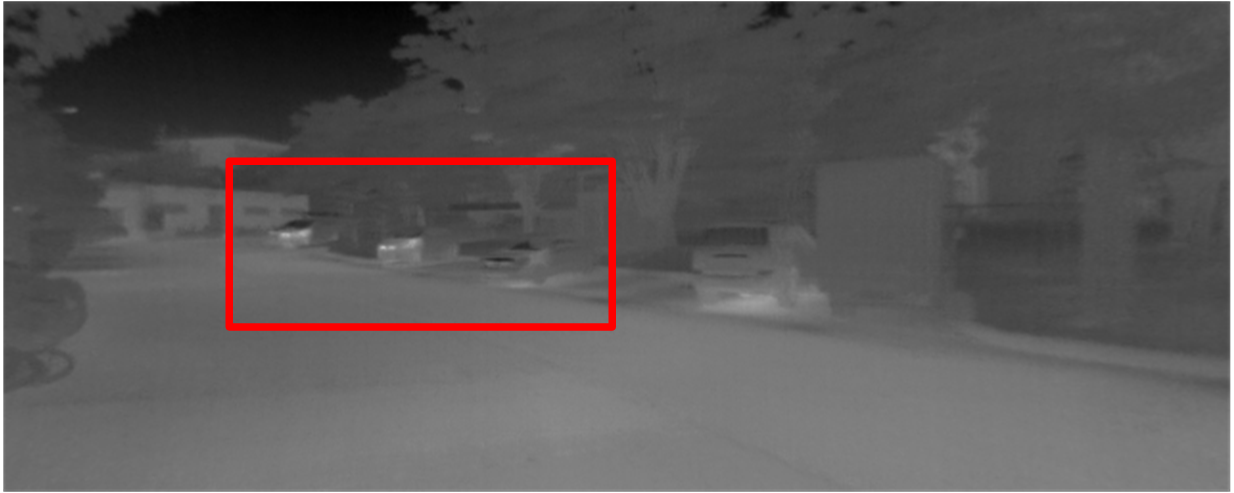}
\includegraphics[width=0.33\linewidth,height=1.8cm]{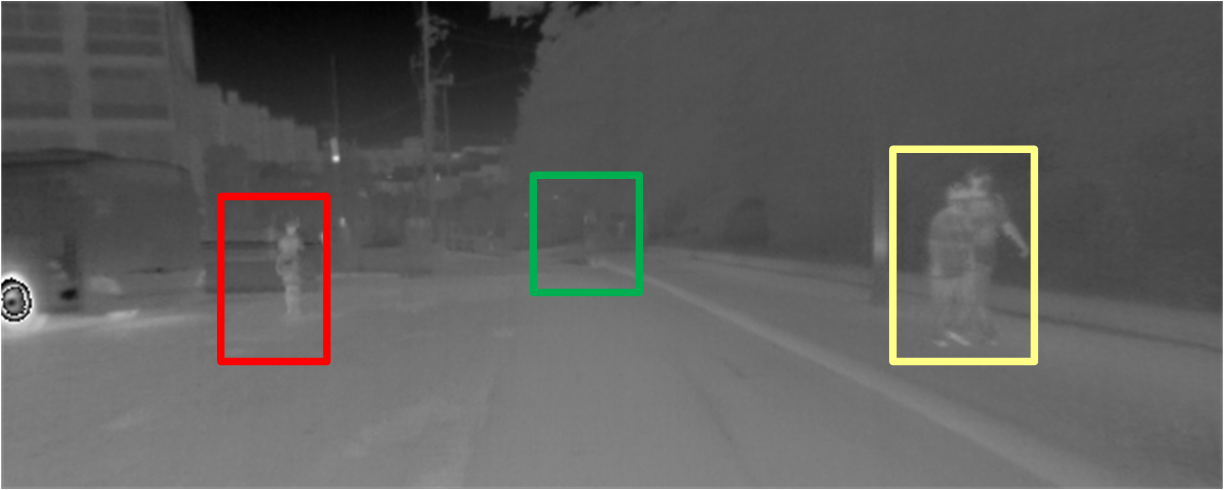}
}\\
\vspace{-1em}
\subfloat{
\parbox[t]{2mm}{\rotatebox{90}{\small ~~~~~~~~~(c)}}
\includegraphics[width=0.33\linewidth]{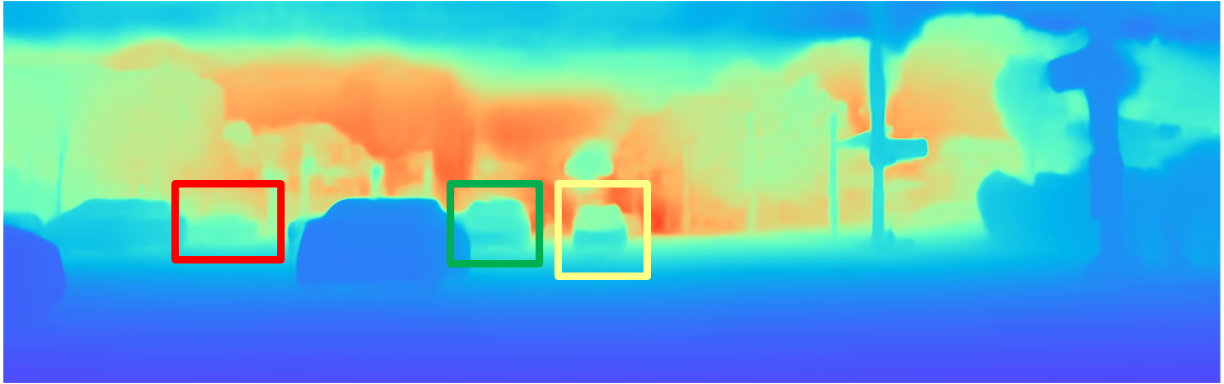}
\includegraphics[width=0.33\linewidth]{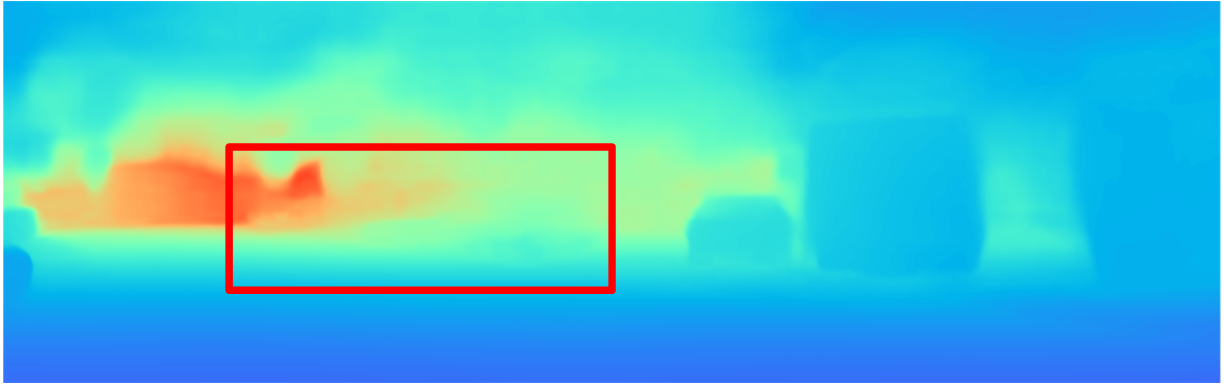}
\includegraphics[width=0.33\linewidth]{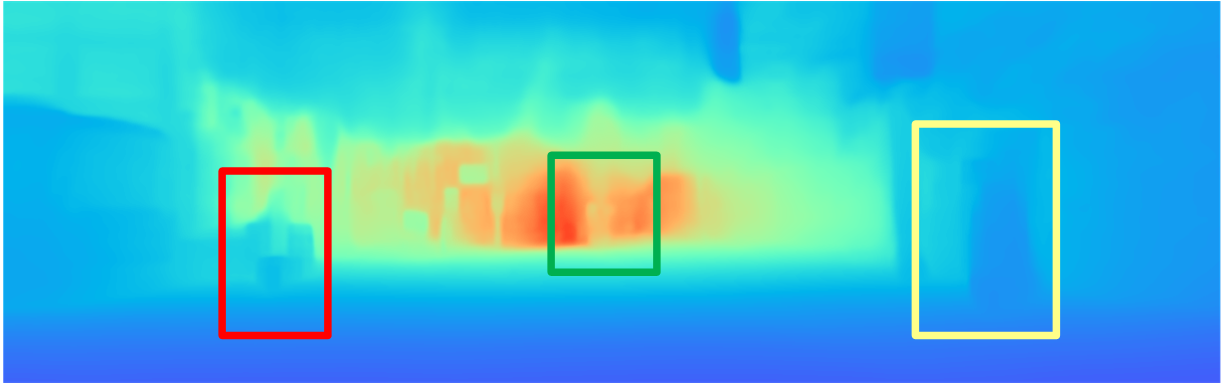}
}\\
\vspace{-1em}
\subfloat{
\parbox[t]{2mm}{\rotatebox{90}{\small ~~~~~~~~~(d)}}
\includegraphics[width=0.33\linewidth]{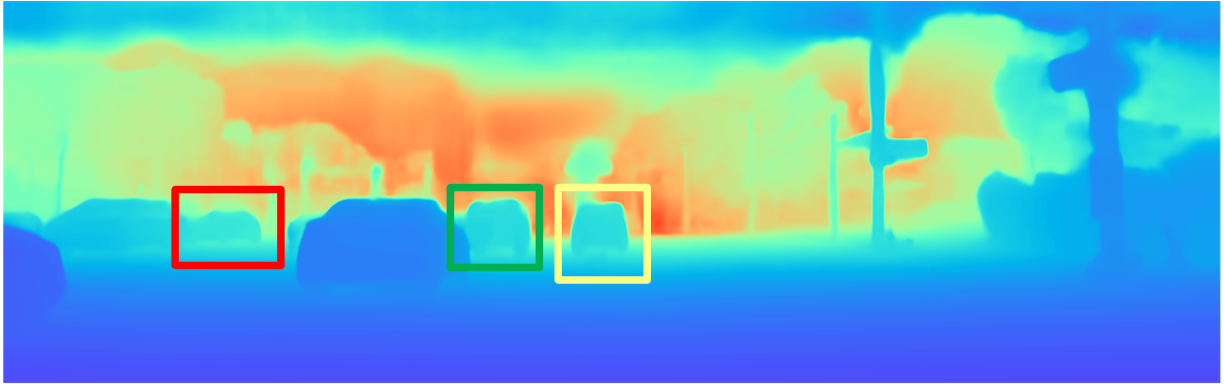}
\includegraphics[width=0.33\linewidth]{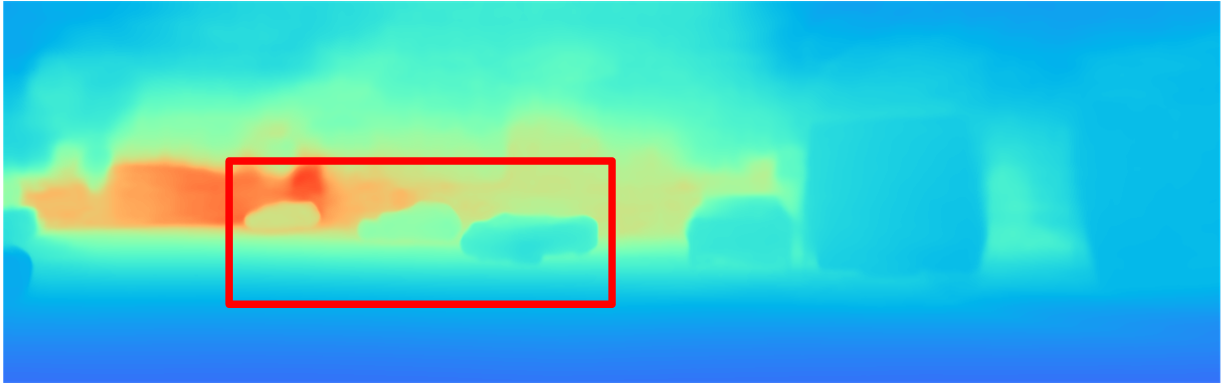}
\includegraphics[width=0.33\linewidth]{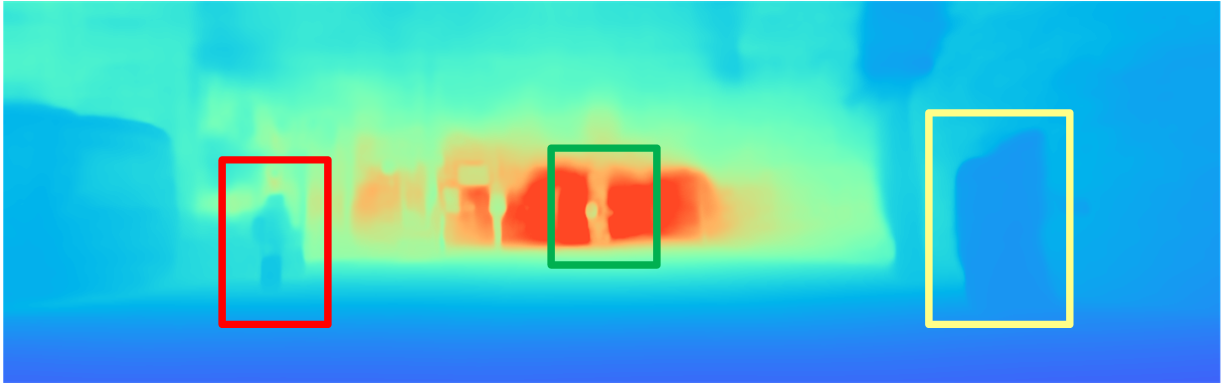}
}\\

\caption{\protect\textbf{Qualitative results of our method on the MS$^2$ dataset~\protect\cite{shin2023deep}.} (a) RGB (b) THR (c) Predicted depth from RGB image only (d) Predicted depth by our method.
}

\label{fig:ms2-results}
\vspace{-0.4cm}
\end{figure*}

\section{Experiments}
\label{sec:exp}
\subsection{Datasets and Experimental Settings}
\noindent\textbf{MS$^2$ dataset~\cite{shin2023deep}}. 
Multi-Spectral Stereo (MS$^2$) Outdoor Driving Dataset is a large-scale multi-modal dataset, which including stereo RGB, stereo NIR, stereo thermal and stereo LiDAR along with GNSS/IMU data. In our experiment, we only use the left camera of the stereo RGB and thermal cameras. 
MS$^2$ dataset contains about 195K synchronized data pairs, which are split in  three test scenarios: Day, Night and Rain.
 
\noindent\textbf{ViViD++ \cite{lee2022vivid++}}. 
Vision for Visibility Dataset (ViViD++) is the first dataset to enclose information from multiple types of aligned alternative vision sensors, which is obtained from poor lighting conditions. ViViD++ provides normal and poor illumination sequences recorded by thermal, depth, and temporal difference sensor. In our experiment, we only use indoor data to evaluate our methods. 

\noindent\textbf{Benchmarks.}
To the best of our knowledge, there is currently no study of predicting depth using both RGB and thermal images simultaneously.  In this context, 
we adopt two state-of-the-art multi-modal models MURF~\cite{xu2023murf} and MCT~\cite{wang2023mct} for comparison. We add depth prediction head on the model and train them with depth ground-truth for a fair comparison.
Results of SOTA monocular depth estimation methods using one modal are also compared.

\subsection{Evaluations}
\noindent\textbf{Evaluation on MS$^2$.} For outdoor scenes, we evaluate our method on the MS$^2$ dataset. As shown in Tab.~\ref{tab:result-ms2}, we first show the results of existing SOTA monocular depth estimation methods employing either RGB or thermal images as a reference. 
Utilizing both RGB and thermal images, we can see that our method outperforms other methods by a significant margin. Specifically, the ``Abs-Rel'' errors are decreased by 10.2\% compared to MURF, and by 9.4\% compared to MCT. 
There is also a significant advantage in ``$\delta <$ 1.25$^i$'' compared to MCT and MURF.

To further probe the robustness, we asses the performance in each test subset (day, night, rain) in Tab.~\ref{tab:different weather}.
Evidently, across each scenario, our method employing both RGB and thermal modalities consistently outperforms the use of a single modality. 
This substantiates that our approach is not merely selecting the most effective modality, but rather effectively fusing multi-modal information to yield accurate geometric predictions.
As shown in Fig.~\ref{fig:first_image} and Fig.~\ref{fig:ms2-results}, in scenes such as dark nights and rainy days that are unfavorable for RGB, fusion of thermal images can significantly improve depth prediction.

\noindent\textbf{Evaluations on ViViD++.}
For indoor dark scenes, we evaluate our method on the ViViD++ dataset. As shown in Tab.~\ref{tab:result-ViViD++}, our method performs significantly better than other multi-modal methods MCT and MURF. Specifically, the ``Abs Rel'' error of our method reduces from 0.062 to 0.054 (14.8\%) compared to the MCT, and the accuracy $\delta<1.25^i$ is significantly higher than other methods. 

\subsection{Limitation of Existing Multi-modal Fusion}

Due to the lack of multi-modal models tailored for depth estimation, we adopt two state-of-the-art multi-modal fusion methods MCT~\cite{wang2023mct} and MURF~\cite{xu2023murf}, whose inputs are RGB and thermal images. At the network output layer, we replace it with the task head suitable for deep prediction and use depth ground-truth for supervision. 
However, as shown in Tab.~\ref{tab:result-ms2} and~\ref{tab:result-ViViD++}, compared with single mode, the improvement of multi-modal methods MURF and MCT is very limited.
The reasons are two-fold. Firstly, the existing multi-modal methods do not take into account the geometric constraints under different perspectives in 3D vision. High-value parameters such as intrinsic and extrinsic parameters cannot participate in network calculations, which results in depth misalignment from different modalities.
Secondly, the existing multi-modal methods cannot identify the depth-related advantages of each modality effectively. It is inefficient to input all the information into the fusion network, and the worthless features can mislead the fusion network.


\subsection{Ablation Study}

\noindent\textbf{3D Cross-modal Transformation.}
As illustrated in Tab.~\ref{tab:ablation cmt and discriminator}, 3D cross-modal transformation (CMT)  plays a critical role in the cross-modal depth estimation task, which can significantly improve the performance of multi-modal fusion. This enhancement stems from the alignment of information derived from both modalities.

\noindent\textbf{Confidence Predictor Network.}
As shown in Tab.~\ref{tab:ablation cmt and discriminator}, the confidence predictor network significantly enhances the effectiveness of multi-modal networks. RGB and thermal cameras have their own advantages and disadvantages in different scenarios, and the confidence predictor network can identify which mode's predicted depth is closer to the ground-truth, providing effective guidance information for subsequent fusion networks.

\noindent\textbf{Loss Function.}
Tab.~\ref{tab:ablation loss} illustrates the outcomes of training networks through diverse combinations of loss function terms. $\mathcal{L}_{coa}$ guarantees the precision of coarse depths (\emph{i.e.}, $\mathbf{D}_{RGB}$ and $\mathbf{D}_{THR}$), which is crucial for ensuring  3D cross-modal transformation.
On the other hand, $\mathcal{L}_{con}$ enables the confidence predictor network to identify the advantages of each modality, and compute the corresponding confidence map. Without $\mathcal{L}_{con}$, the network can implicitly learn the relative correct relationship between two modalities. However, it is less effective than using the direct loss function.

\begin{figure}[t!]
\centering

\subfloat{
\includegraphics[width=\linewidth]{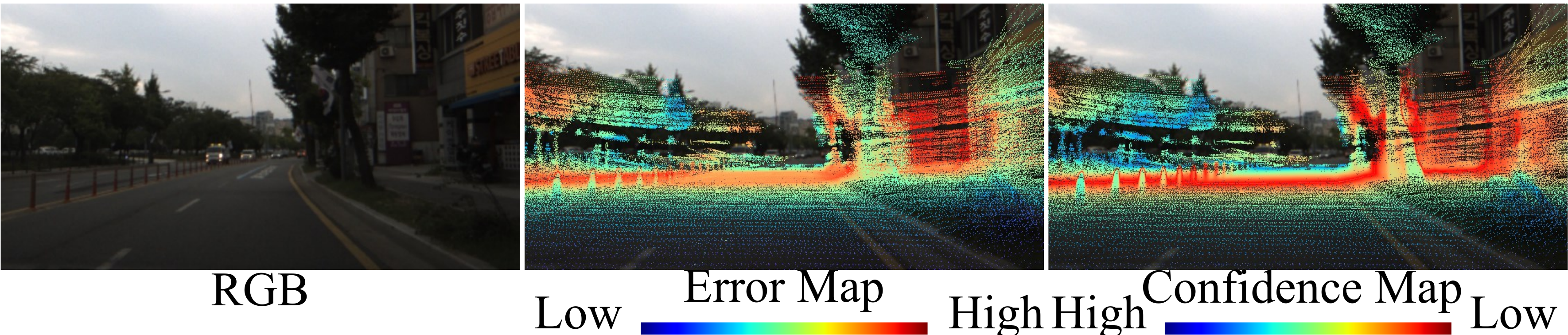}
}
\vspace{-0.5em}
\caption{\textbf{Visualization of depth error map of RGB and the corresponding confidence map. 
} 
The error map is computed between the ground-truth and predicted depth map. The predicted confidence map reflects the dominant region of each modality.
}
\label{fig:confidence map}
\end{figure}

\subsection{Visualization of Confidence Maps}
As shown in Fig.~\ref{fig:confidence map}, we visualize the depth error map of RGB  and predicted confidence map. Error map 
The confidence map can accurately reflect the error of RGB depth estimation, and the high confidence region is consistent with the low error region. This shows that the confidence predictor network driven by confidence loss can accurately judge the dominant region and high-value information of each modality, which gives important guidance to the subsequent fusion network.
\begin{table}[t!]
\begin{center}
\renewcommand\arraystretch{0.9}
\setlength\tabcolsep{8.0pt} 
\small
\begin{tabular}{cc|cc|c}
\hline
CMT
&CPN
&AbsREL $\downarrow$
&RMSE$\downarrow$
&$\delta<1.25\uparrow$\\
\hline
\hline
\checkmark &  &                        0.119 & 5.121 & 0.865 \\
  & \checkmark &                      0.110 & 5.052 & 0.878\\
\checkmark  & \checkmark  & \textbf{0.106} & \textbf{5.003} & \textbf{0.882}\\
\hline

\end{tabular}
\vspace{-0.5em}
\caption{\label{tab:ablation cmt and discriminator} The ablation study for 3D cross-modal transformation (CMT) and confidence predictor network (CPN). } 
\end{center}
\vspace{-0.5em}
\end{table}

\begin{table}[t!]
\begin{center}
\renewcommand\arraystretch{0.9}
\setlength\tabcolsep{5.0pt} 
\small
\begin{tabular}{ccc|cc|c}

  \hline
$\mathcal{L}_{coa}$
& $\mathcal{L}_{con}$ 
& $\mathcal{L}_{SILog}$ 
&AbsREL $\downarrow$
&RMSE$\downarrow$
&$\delta<1.25\uparrow$\\
\hline
  \hline
 &  & \checkmark                                & 0.130 & 5.502 & 0.839 \\
\checkmark &  & \checkmark                      & 0.121 & 5.384 & 0.853 \\
  & \checkmark & \checkmark                     & 0.117 & 5.129 & 0.874\\
\checkmark  & \checkmark & \checkmark & \textbf{0.106} & \textbf{5.003} & \textbf{0.882}\\
\hline

\end{tabular}
\vspace{-0.5em}
\caption{\label{tab:ablation loss} The ablation study for loss function.} 
\end{center}
\vspace{-1em}
\end{table}

\section{Conclusion}
In this paper, we presented a novel approach to robust monocular depth estimation via cross-modal fusion of RGB and thermal images. Our method leverages the complementary advantages of both modalities to improve the accuracy and robustness of depth estimation in challenging environments such as night, rain, and low-light indoor scenes. 
We designed a confidence predictor network and a fusion network to identify and fuse the advantages of each modality. We have demonstrated the effectiveness of our approach through extensive experiments on several benchmark datasets, showing significant improvements over state-of-the-art methods. 


\bibliographystyle{named}
\bibliography{ijcai24}

\end{document}